\documentclass[twoside,11pt]{article}

\usepackage{jmlr2e}
\usepackage{epstopdf}

\begin{document}

\title{LOFS: Library of Online Streaming Feature Selection}

\author{\name Kui Yu \email Kui.Yu@unisa.edu.au \\
       \addr School of Information Technology and Mathematical Sciences\\
       University of South Australia\\
       Adelaide, 5095, SA, Australia
       \AND
       \name Wei Ding \email ding@cs.umb.edu\\
       \addr Department of Computer Science\\
         University of Massachusetts Boston\\
         Boston, 02125-3393, MA  USA
       \AND
       \name Xindong Wu \email xwu@uvm.edu.cn\\
       \addr Department of Computer Science\\
        University of Vermont\\
        Burlington, 05405, VT, USA}

\editor{}

\maketitle

\begin{abstract}%   <- trailing '%' for backward compatibility of .sty file
As an emerging research direction, online streaming feature selection deals with sequentially added dimensions in a feature space while the number of data instances is fixed. Online streaming feature selection provides a new, complementary algorithmic methodology to enrich online feature selection, especially targets to high dimensionality in big data analytics. This paper introduces the first comprehensive open-source library for use in MATLAB that implements the state-of-the-art algorithms of online streaming feature selection. The library is designed to facilitate the development of new algorithms in this exciting research direction and make comparisons between the new methods and existing ones available. LOFS is available from https://github.com/kuiy/LOFS.
\end{abstract}


\section{Introduction}

Feature selection is to select a parsimonious feature subset to improve model interpretability and efficiency without degrading model accuracy~\citep{guyon2003introduction}. Traditionally, online feature selection deals with the observations sequentially added while the total dimensionality is fixed~\citep{wang2014online,hoi2014libol}. However, in many real world applications, it is either impossible to acquire an entire feature set or impractical to wait for the complete set before feature selection starts. For instance, in Twitter, trending topics keep changing over time, and thus the dimensionality of data is changed dynamically. When a new top topic appears, it may come with a set of new keywords, which usually serve as key features to identify new hot topics.  In bioinformatics, it is expensive for feature selection to acquire an entire set of features for each data observation due to the high cost in conducting wet lab experiments~\citep{wang2014online}. Many big data applications call for online streaming feature selection to consume sequentially added dimensions over time.

In general, assuming $S$ is the feature set containing all features available till time $t_{i-1}$, and $C$ is the class attribute, then a training data set $D$ is defined by $D=\{S,C\}$, which is a sequence of features that is presented over time. As we process one dimension at a time, the research problem is that at any time $t_i$, how to online maintain a minimum size of feature subset $S{_{t_i}^{\star}}$ of maximizing its predictive performance for classification. If $F_i$ is a new coming feature at time $t_i$, $S{_{t_{i-1}}^{\star}}\subset S$ is the selected feature set till time $t_{i-1}$ and $P(C|\zeta)$ denotes the posterior probability of $C$ conditioned on a subset $\zeta$, the problem of online streaming feature selection is formulated as Eq.(1).

\begin{equation}
 S_{t_i}^{\star}  =  \arg\min_{S'}\{|S^{\prime}|:S^{\prime}=\mathop{\arg\max}_{\zeta\subseteq \{S{_{t_{i-1}}^{\star}}\cup F_i\}}P(C|\zeta)\}.
\end{equation}

To solve Eq.(1), currently the state-of-the-art algorithms include Grafting~\citep{perkins2003online}, Alpha-investing~\citep{zhou2006streamwise} , OSFS~\citep{wu2010online}, Fast-OSFS~\citep{wu2013online}, and SAOLA~\citep{yu2014towards}. All of those algorithms only deal with one dimension at a time upon its arrival.

Group information sometimes is embedded in a feature space. For instance, in image analysis, features are generated in groups which represent color, texture and other visual information. If  $G_{t_{i-1}}$ is the set of all feature groups available till time $t_{i-1}$, then at at time $t_{i-1}$, $D$ is denoted by $D=\{G_{t_{i-1}}, C\}$, which is a sequence of feature groups that is added sequentially. To consume grouped features sequentially added over time, online selection of dynamic groups is formulated as Eq.(2).

\begin{equation}
\begin{array}{ll}
 &G_{t_i}^{\star}= \mathop{\arg\max}_{G_{\zeta}\subseteq \{G_{t_{i-1}}^{\star}\cup G_i\}}P(C|G_\zeta)\\
 &s.t.\\
 &\ (a)\forall F_k\in G_j, G_j\subset G_{t_i}^{\star}, P(C|\{G_j-\{F_k\},F_k\})\neq P(C|\{G_j-\{F_k\}\}) \\
 &\ (b) \forall G_j\subset G_{t_i}^{\star}, P(C|\{G_{t_i}^{\star}-G_j, G_j\})\neq P(C|\{G_{t_i}^{\star}-G_j\}) .
\end{array}
\end{equation}

In Eq.(2), $G_i$ is a new coming group at time $t_i$, and $G_{t_{i-1}}^{\star}\subset G_{t_{i-1}}$ is the set of selected groups till time $t_{i-1}$. Eq.(2) attempts to yield a set of groups at time $t_i$ that is as parsimonious as possible at the levels of both intra-groups (constraint (a)) and  inter-groups (constraint (b)) simultaneously for maximizing its predictive performance for classification. To online utilize grouped features, the group-SAOLA algorithm was proposed~\citep{Yu2015online}.

As an emerging research direction, online streaming feature selection provides a new, complementary algorithmic methodology to enrich online feature selection, especially addresses high dimensionality in big data analytics. But to the best of our knowledge, there is no comprehensive open-source packages existing for this problem. To facilitate research efforts on this novel direction, we develop the open-source library called LOFS (\textbf{L}ibrary of \textbf{O}nline streaming \textbf{F}eature \textbf{S}election).

The main contribution of LOFS lies on three aspects. (1) It is the first comprehensive open-source library for implementing algorithms of online streaming feature selection. (2) It provides the state-of-the-art algorithms of online streaming feature selection mainly developed by our research group, including the algorithms to learn features added individually over time and the methods to mine grouped features added sequentially. (3) It is written in MATLAB, easy to use, and completely open source. We hope it will facilitate the development of new online algorithms for tackling the grand challenges of high dimensionality in big data analytics, and encourage researchers to extend LOFS and share their algorithms through the LOFS framework.

\section{Architecture of LOFS}

The LOFS architecture is based on a separation of three modules, that is, CM (\textbf{C}orrelation \textbf{M}easure), Learning, and SC (\textbf{S}tatistical \textbf{C}omparison), as shown in Figure 1. The learning module consists of two submodules, LFI (\textbf{L}earning \textbf{F}eatures added \textbf{I}ndividually) and LGF (\textbf{L}earning \textbf{G}rouped \textbf{F}eatures added sequentially).
\begin{figure}
\centering
\includegraphics [height=2.5in,width=5in]{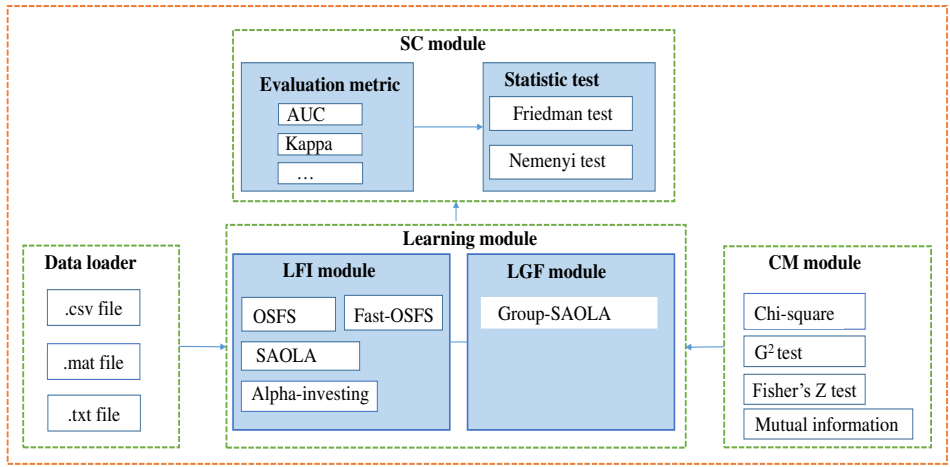}
\caption{Architecture of LOFS}
\end{figure}

In the CM module, the library provides four measures to calculate correlations between features, Chi-square test, $G^2$ test, the Fisher's Z test, and mutual information, where Chi-square test, $G^2$ test, and mutual information for dealing with discrete data while the Fisher's Z test for handling continuous data.

With the measures above, the LFI module includes Alpha-investing~\citep{zhou2006streamwise}, OSFS~\citep{wu2010online}, Fast-OSFS~\citep{wu2013online}, and SAOLA~\citep{yu2014towards} to learn features added individually over time, while the LGF module provides the group-SAOLA algorithm~\citep{Yu2015online} to online mine grouped features added sequentially.

Based on the learning module, the SC module provides a series of performance evaluation metrics (i.e., prediction accuracy, AUC, kappa statistic, and compactness, etc ). To conduct statistical comparisons of algorithms over multiple data sets, the SC model further provides the Friedman test and the Nemenyi test~\citep{demvsar2006statistical,li2014sparse}.

The three modules in the LOFS architecture are designed  independently, and all codes follow the MATALB standards. This guarantees that the LOFS library is simple, easy to implement, and easily extendable. One can easily add a new algorithm to the LOFS library and share it through the LOFS framework.

\section{Usage of LOFS}
The LOFS library comes with detailed documentation. The documentation is available from https://github.com/kuiy/LOFS. This documentation describes the setup and usage of LOFS. All the functions and related data structures are explained in detail. In Figure 2, we give an example to illustrate how to implement the functions of the Fast-OSFS algorithm in LOFS, such as loading data, setting parameters, running algorithms, and evaluating performance.
\begin{figure}
\centering
\includegraphics [height=2.7in,width=5in]{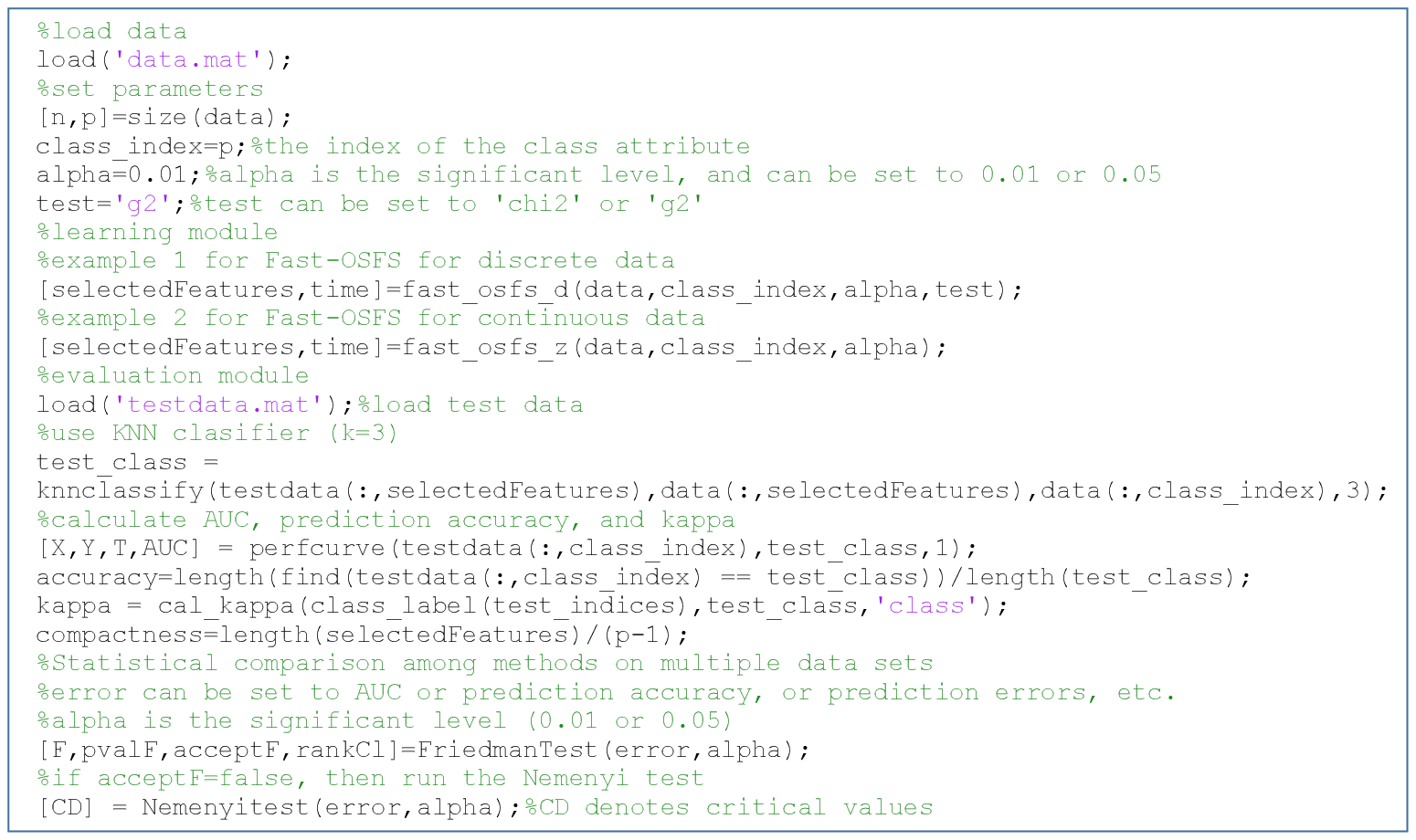}
\caption{Example of implementing LOFS}
\end{figure}

In addition to the documentation, the extensive performance comparisons between online streaming feature selection and traditional online feature selection, and the empirical comparisons between online streaming feature selection and the traditional feature selection, such as GDM~\citep{zhai2012discovering}, FCBF~\citep{yu2004efficient}, and SPSF-LAR~\citep{zhao2013similarity},  can be found in Yu et al.~\citep{Yu2015online}.

\section{Conclusion and Future Work}
This paper presents LOSF, an easy-to-use open-source package for online streaming feature selection to facilitate  research efforts in machine learning and data mining. It is the first open-source package for online streaming feature selection. Through the LOFS framework, we hope that it will facilitate researchers to develop new online learning algorithms for big data analytics and share their algorithms. Future versions of LOFS will have the ability to load and save the ARFF (Weka Attribute-Relation File Format) files and LIBSVM data files\footnote{https://www.csie.ntu.edu.tw/$\sim$cjlin/libsvmtools/datasets/}. LOFS will keep trying its best to integrate new developed algorithms in future.

% Acknowledgements should go at the end, before appendices and references

%\acks{This work is partly supported by the Program for Changjiang Scholars and Innovative %Research Team in University (PCSIRT) of the Ministry of Education, China (under grant IRT13059), %and the National Natural Science Foundation of China (under grants 61305064 and 61572492). }
%the National 973 Program of China (under grant 2013CB329604)

%\vskip 0.2in
\bibliography{lofs}

\end{document}